\documentclass[pdflatex,sn-mathphys-num]{sn-jnl}

\usepackage{graphicx}%
\usepackage{multirow}%
\usepackage{amsmath,amssymb,amsfonts}%
\usepackage{amsthm}%
\usepackage{mathrsfs}%
\usepackage[title]{appendix}%
\usepackage{xcolor}%
\usepackage{textcomp}%
\usepackage{manyfoot}%
\usepackage{booktabs}%
\usepackage{algorithm}%
\usepackage{algorithmicx}%
\usepackage{algpseudocode}%
\usepackage{listings}%

\usepackage{array}
\usepackage{stfloats}
\usepackage{url}
\usepackage{verbatim}
\usepackage{gensymb}
\usepackage{dirtytalk}
\usepackage{tabularray}
\usepackage{booktabs}
\usepackage{color}

\usepackage{amsfonts}
\usepackage{amsmath,amssymb,amsfonts}
\usepackage{amsthm}
\usepackage{appendix}[title]
\usepackage{algorithm}
\usepackage{algorithmicx}
\usepackage{algpseudocode}
\usepackage{booktabs}
\usepackage{color}
\usepackage{fancyhdr}
\usepackage{gensymb}
\usepackage{graphicx}
\usepackage{hhline}
\usepackage{hyperref}
\usepackage{listings}
\usepackage{manyfoot}
\usepackage{mathrsfs}
\usepackage{multirow}
\usepackage{optidef}
\usepackage{outlines}
\usepackage{setspace}
\usepackage{siunitx}
\usepackage{textcomp}
\usepackage{xcolor}
\usepackage{threeparttable}

\definecolor{burntorange}{rgb}{0.8, 0.33, 0.0}
\definecolor{CFCblue}{RGB}{3, 70, 148}
\definecolor{Magenta (process)}{rgb}{1.0, 0.0, 0.56}

\theoremstyle{thmstyleone}%

\theoremstyle{thmstyletwo}%

\theoremstyle{thmstylethree}%

\begin{document}

\title[Article Title]{An MRI-Guided Robotic System to Improve Hippocampal Access for Epilepsy Interventions}


\author*[1]{\fnm{John E.} \sur{Peters}}\email{john.e.peters.1@vanderbilt.edu}
\equalcont{These authors contributed equally to this work.}

\author*[1]{\fnm{Abby M.} \sur{Grillo}}\email{abby.m.grillo@vanderbilt.edu}
\equalcont{These authors contributed equally to this work.}

\author[1]{\fnm{Mahshid} \sur{Mansouri}}

\author[1]{\fnm{Daniel S.} \sur{Esser}}

\author[2]{\fnm{Sarah} \sur{Garrow}}

\author[1]{\fnm{Nithin S.} \sur{Kumar}}

\author[3]{\fnm{Dario J.} \sur{Englot}}

\author[4]{\fnm{Joseph S.} \sur{Neimat}}

\author[2]{\fnm{William A.} \sur{Grissom}}

\author[1,3]{\fnm{Robert J.} \sur{Webster III}}

\author[1,3]{\fnm{Eric J.} \sur{Barth}}

\affil[1]{\orgdiv{Department of Mechanical Engineering}, \orgname{Vanderbilt University}, \orgaddress{\city{Nashville}, \state{TN}, \country{USA}}}

\affil[2]{\orgdiv{Department of Biomedical Engineering}, \orgname{Case Western Reserve University}, \orgaddress{\city{Cleveland}, \state{OH}, \country{USA}}}

\affil[3]{\orgdiv{Department of Neurological Surgery}, \orgname{Vanderbilt University Medical Center}, \orgaddress{\city{Nashville}, \state{TN}, \country{USA}}}

\affil[4]{\orgdiv{Department of Neurological Surgery}, \orgname{University of Louisville Health}, \orgaddress{\city{Louisville}, \state{KY}, \country{USA}}}


\abstract{This paper presents an MRI-guided robotic system that improves hippocampal access by delivering a curved needle-like laser ablator through the foramen ovale, a natural opening in the base of the skull. Both the delivery path and the curved nature of the needle improve upon current clinical straight-line laser interstitial thermal therapy (LITT) as measured by hippocampal cannulation percentage. We describe the design of the robotic system which includes positioning, aiming, and curved needle deployment stages, followed by experimental results assessing cannulation percentages using MRI images in phantoms. In three curvilinear, single-insertion experiments, we achieved cannulation percentages of 93.5\%, 96.5\%, and 60.4\%, exceeding reported clinical averages of 50-60\%. Seizure control is believed by physicians to be a function of hippocampal volume treated, and our system provides a means of treating a greater volume of the hippocampus with LITT-based interventions.}


\maketitle

\section{Introduction}\label{intro}
Epilepsy affects 68 million people worldwide, causing debilitating seizures that not only impair quality of life, but can be fatal \cite{nevalainen_epilepsy-related_2014, Wiebe1999, Ficker1998}. While antiepileptic drugs can reduce seizure frequency, 40\% of the patient population is drug-resistant \cite{kwan2000epilepsy}. The most common drug-resistant subtype, mesial temporal lobe epilepsy, originates in the hippocampus~\cite{thom_review_2014}. 
Surgical removal of the hippocampus (called selective amygdalohippocampectomy) can achieve up to 70\% seizure freedom, but is underutilized due to patient reluctance to undergo invasive open brain surgery, even though surgeons argue that the rewards outweigh the costs \cite{hori2004subtemporal, lutz2004neuropsychological, wieser2003long, uijl2012epilepsy, DeFlon2010empirical}. 

These factors have motivated the development of MRI-guided laser interstitial thermal therapy (LITT) as a minimally invasive alternative. Currently, LITT is delivered along straight trajectories that fail to conform to the natural curvature of the hippocampus. This limits the distance the laser probe can travel through the hippocampus, which is clinically measured as hippocampal cannulation percentage. Surgeons seek to maximize this percentage to increase the quantity of target tissue ablated \cite{wu2015effects}. 

Improved cannulation is clinically important because seizure outcomes depend on head-to-tail hippocampal coverage, which straight-line LITT trajectories often under cover. Insufficient ablation of the head has been linked to persistent seizures \cite{jermakowicz2017laser,Wu2019}, while recent evidence suggests that the tail is an important interventional target in improving epilepsy outcomes \cite{jamiolkowski2024fasciola}. Current strategies to increase coverage include multiple straight trajectories delivered via multiple burr holes, but these add operative time and increase collateral damage to healthy tissue \cite{liu2021two}. In this paper, we propose a curvilinear approach designed to access both the head and tail in a single trajectory.
While cannulation does not directly determine the final ablation volume in LITT, intrahippocampal placement is desirable because it centers the heat source within the target tissue, reducing reliance on thermal conduction from extrahippocampal positions and limiting off-target heating. In clinical practice, incomplete cannulation is often compensated by thermal spread, with laser probes achieving radial penetration depths of up to 13 mm  \cite{wu2015effects,ahrar2010preclinical}. 

The cannulation percentage clinically for LITT is 50-60\% \cite{wu2014extraventricular, wu2015effects,vakharia_automated_2018,kang2016laser}, and the resulting limited head-to-tail coverage has been associated with re-operations and poor seizure control \cite{jermakowicz2017laser,Wu2019,jamiolkowski2024fasciola}. As such, LITT seizure freedom rates remain 16\% lower than anterior temporal lobectomy and selective amygdalohippocampectomy \cite{jermakowicz2017laser, liu2021two, gupta2020robot, wu2014extraventricular, wu2015effects, Wu2019, jamiolkowski2024fasciola, Willie2014}. In this work, we aim to increase hippocampal cannulation by robotically delivering the LITT probe through a single, curvilinear, anatomy-conforming trajectory -- i.e.\ a trajectory that has the potential to enable head-to-tail traversal and eventually increase overall ablation coverage.

MR-conditional robots have been the subject of much study since Masamune et al.\ \cite{masamune1995development} introduced the first one in the 1990s. Since then, substantial research has established that needle-placement robots for stereotactic neurosurgery can be safely integrated into the MRI environment and achieve clinically relevant accuracy \cite{cole2009design,li2014robotic,li2020fully,patel2020integrated,li_jinhua_2024mri_semiEnclosed}. In-bore, MR-conditional robots have been implemented using several non-magnetic actuation approaches, including piezoelectric motors \cite{tsekos2005prototype,tsekos2007magnetic,fischer2008mri,krieger2011development,su2014piezoelectrically}, hydraulic actuation \cite{guo2018compact,qiu2021mri,mendoza2019testbed}, and pneumatic actuation \cite{stoianovici2007new, fischer2008mri_PNEUrobot,comber2014design,stoianovici2016mr,groenhuis2017design,chen2017characterization,groenhuis2018stormram,chen2019mr,gunderman2023non}. However, nearly all prior systems are limited to straight trajectories delivered through burr-hole access. In contrast, our work enables percutaneous, curved trajectories to maximize hippocampal cannulation while avoiding burr holes.

Curved needle paths offer the possibility of generalizing MRI-guided needle placement beyond traditional straight insertions. Nonlinear trajectories in the brain can avoid eloquent regions and enable treatment of geometrically complex targets like the hippocampus. Several robotic approaches have been proposed to achieve curved trajectories, including steerable needle systems \cite{minhas2007modeling, okazawa2005hand} and non-needle continuum robots \cite{bajo2016hybrid, runciman2019soft, esser2025encoding}. One needle-based approach to achieving curved trajectories is use of the concentric tube robot (CTR), a class of steerable needle composed of nested, precurved elastic tubes that generate controllable curved shapes through relative translation and rotation \cite{mahoney2019review}. These robots can be fabricated from superelastic materials such as nitinol and are shaped to achieve high curvatures, enabling access to anatomical regions that are difficult to reach with conventional rigid instruments. CTRs have been widely investigated (see \cite{nwafor_design_2023} for a review) for minimally invasive surgical applications including neurosurgery \cite{rox2020mechatronic}, intracardiac procedures \cite{bedell2011design}, and prostate interventions \cite{hendrick2015hand}, among others. CTRs have also been deployed using MR-conditional robots \cite{su2016concentric}. We previously contributed a novel pneumatic actuation approach, leveraging helical and toroidal bellows that can be additively manufactured as a single unit \cite{Comber2016}.

\begin{figure}[t]
\centering
\includegraphics[width=\columnwidth]{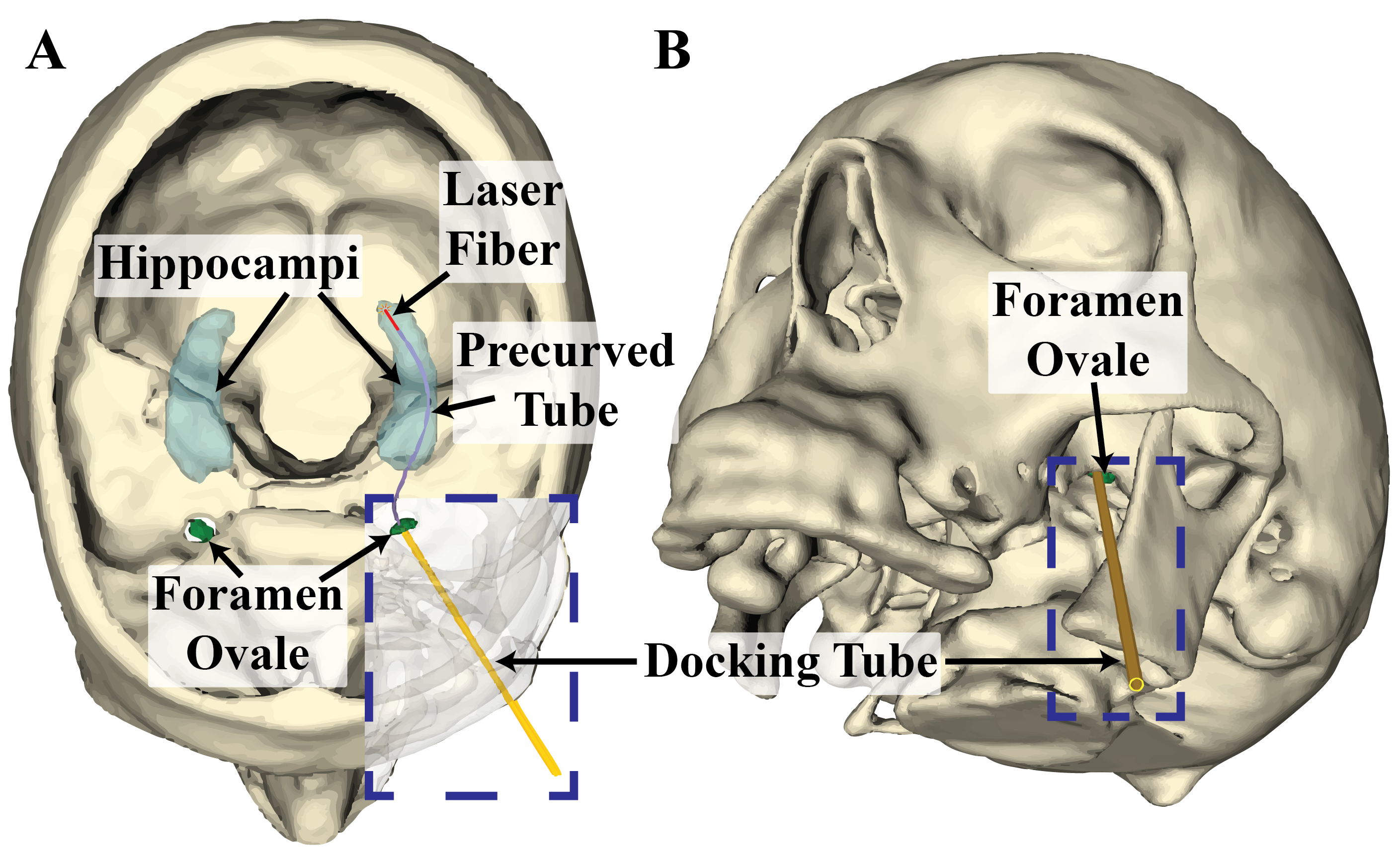}
\caption{Transforaminal ablation approach in a patient skull. (A) Concentric docking tube, precurved tube, and laser fiber traversed through the foramen ovale and hippocampus. (B) Percutaneous approach of the docking tube to the foramen ovale.}
\label{fig:sysConcept}
\end{figure}

Our group initially explored an occipital approach to targeting the hippocampus using a CTR as a steerable needle \cite{gilbert2015concentric}. We demonstrated that helical trajectories can follow in closer proximity to the hippocampal medial axis compared to straight trajectories. Then, to reduce invasiveness, we proposed the transforaminal approach, shown in Fig.\ \ref{fig:sysConcept}, where the CTR is delivered through a percutaneous needle insertion into the patient’s cheek \cite{Comber2017Optimization, Granna2022, Grillo2023} to access a natural opening in the skull base, called the foramen ovale. This concept was inspired by prior clinical use of the foramen ovale for treating trigeminal neuralgia and placing recording electrodes for epilepsy diagnosis \cite{wieser1985foramen, bale2006frameless}, but it had never been used for LITT. In the foramen ovale approach, we proposed using a set of tubes consisting of a straight outer tube inserted up to the foramen ovale (herein referred to as the docking tube) that deploys a helically precurved tube containing a straight optical laser fiber. The precurved tube was designed to follow the medial axis of the hippocampus, delivering laser thermal energy at various planned locations to maximize ablation coverage. We previously addressed the planning problem for this procedure in \cite{Granna2022}. 

In this paper, we advance MRI-guided transforaminal hippocampal access in two important ways. First, we address key limitations of our prior robotic systems, which were limited to 2-degree-of-freedom (DOF) CTR deployment without robotic aiming or positioning stages \cite{comber2015dissertation}. As such, we contribute a 5-DOF MR-conditional robotic system for in-bore, MRI-guided transforaminal aiming and deployment of CTRs. The system includes a redesigned CTR actuation unit, a robotic aiming arm that maintains a remote center-of-motion at the foramen ovale for precise trajectory alignment, and a positioning platform for in-bore positioning. 
Second, we present the first experiments demonstrating closed-loop MRI-guided deployment of concentric tube robots along realistic hippocampal anatomy using MRI feedback. Using a single curved trajectory, the system cannulated an average of 83.5\% of the hippocampus across three phantom experiments (93.5\%, 96.5\%, and 60.4\%), which exceeds the 50-60\% average reported for conventional straight-trajectory clinical approaches \cite{wu2015effects, vakharia_automated_2018, wu2014extraventricular, kang2016laser, liu2021two}. Previous conference papers introduced components of the robotic system, including the pneumatic CTR actuators \cite{Peters2023, PetersHSMR2024}, the positioning platform \cite{grillo_6-dof_2025}, and a curved laser probe \cite{Esser2022}, but they did not demonstrate in-phantom or MRI-guided experiments. Also, the design considerations for MR-conditional robotic access to the brain were characterized \cite{Grillo2023}, but they were not used to design a complete robotic system. Presented here is the first full integration of the complete robotic system, and subsequently the first ever experimental demonstration of MRI-guided robotic aiming and transforaminal CTR deployment through realistic hippocampal anatomy. We report the highest hippocampal cannulation percentage experimentally measured in the literature to date. These results demonstrate the potential of the transforaminal robotic approach to increase hippocampal access and to improve epilepsy treatment.

\section{Materials and Methods}\label{sec2}

\begin{figure*}
  \centering
  \includegraphics[width=\columnwidth]{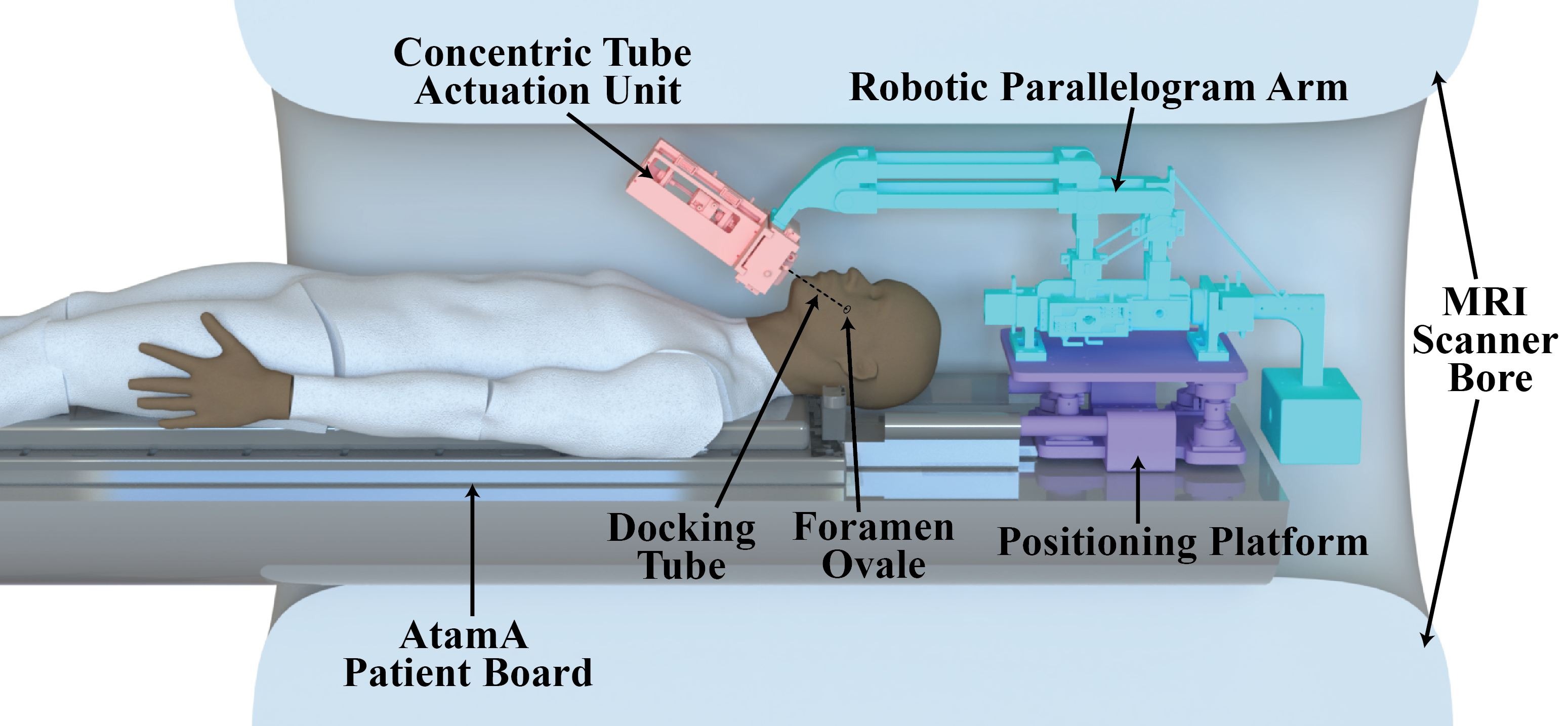}
  \caption{Complete CAD rendering of the entire robotic system depicted for use in the scanner bore with a patient positioned for a percutaneous, transforaminal approach.}
  \label{fig:RoboticSystemOverviewCAD}
\end{figure*}

Our robotic system (Fig.~\ref{fig:RoboticSystemOverviewCAD}) consists of a positioning platform, which is passively adjustable and lockable, and two active stages. These are the Robotic Parallelogram Arm (RPA), which controls the tilt in two directions of the Concentric Tube Actuation Unit (CTAU) and its attached docking tube, and the CTAU itself, which delivers a CTR through the docking tube. The CTR consists of a helically precurved tube with an optical laser fiber nested inside. This MR-conditional, pneumatically actuated robotic system is designed for MRI-guided transforaminal curvilinear LITT of the hippocampus.

\subsection{Clinical Workflow} \label{sec:workflow}
The robotic system is intended to integrate with the existing clinical workflow for LITT while minimizing changes to established transforaminal neurosurgical procedures. After preoperative imaging and selection of the appropriate helical needle (either as a patient-specific needle \cite{Comber2017Optimization} or from a predefined set \cite{Granna2022}), the neurosurgeon manually inserts the docking tube through the patient's cheek up to the foramen ovale under fluoroscopic guidance using standard neurosurgical technique. The patient's head is then immobilized on the AtamA\textsuperscript{\textregistered} transfer board (Monteris Medical), routinely used in clinical LITT, with the robot mounted to the same board to maintain the robot-patient relationship during transport to the MRI scanner. The passive positioning platform is manually adjusted until the Robotic Parallelogram Arm can grasp the docking tube, after which the platform is locked. The patient, transfer board, and attached robot are subsequently transferred together into the MRI scanner, where the Robotic Parallelogram Arm is actuated and controlled to establish the desired docking tube orientation (i.e. aiming) while maintaining an RCM at the foramen ovale, followed by MRI-guided deployment of the CTR and subsequent thermal ablations.

\subsection{Concentric Tube Actuation Unit (CTAU)}
\begin{figure}[!t]
\centering
\includegraphics[width=\columnwidth]{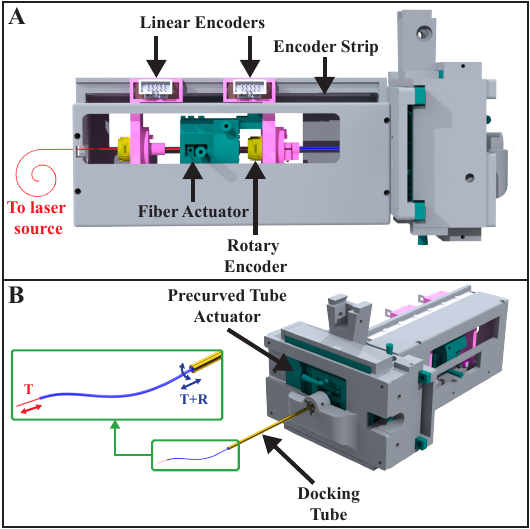}
\caption{The actuation unit of the concentric tube robot from different perspectives. (A) Side view (docking tube hidden for simplicity) detailing the tubes, encoders, and fiber actuator. (B) Isometric view illustrating the concentric tubes, each capable of translation and rotation, as well as the precurved tube actuator. Docking tube is rigidly fixed to the CTAU housing. Inset shows the degrees of freedom (T: translation, R: rotation) associated with the fiber (red) and precurved tube (blue). 
}
\label{fig:CTAUfig}
\end{figure}

\begin{figure}[!t]
\centering
\includegraphics[width=\columnwidth]{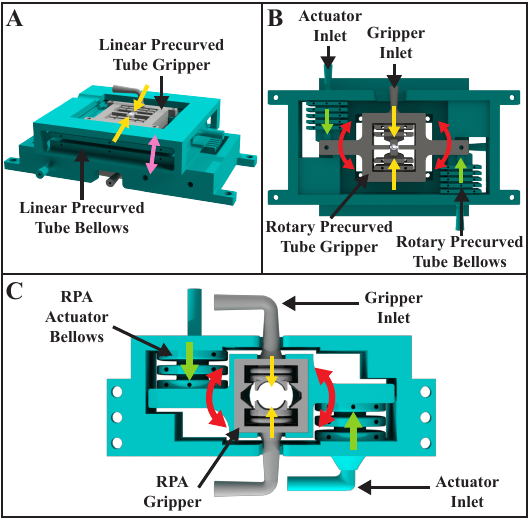}
\caption{Detailed view of the robot's 3D-printed, pneumatic, stepper actuators. (A) Isometric view of the precurved tube actuator (cyan) highlighting the translational DOF (pink arrow) including a linear gripper (grey) and its deflection (yellow arrows). (B) Rear view of the precurved tube actuator detailing the rotational DOF (red arrows) including a rotary gripper (grey) and its deflection (yellow arrows), a pair of rotary bellows and their deflection (green arrows), and various air inlets. (C) Rear view of the RPA actuator (cyan) detailing the rotational DOF (red arrows) including a rotary gripper (grey) and its deflection (yellow arrows), a pair of rotary bellows and their deflection (green arrows), and various air inlets. Each RPA actuator includes a second gripper (not shown) positioned on the opposing end of a common driveshaft. All actuators operate in a stepwise fashion executing either a translational or rotational step using alternating gripper states. Additional detail on the stepwise operation can be found in \cite{Peters2023}.
}
\label{fig:ActuatorsFig}
\end{figure}

The Concentric Tube Actuation Unit (CTAU) at the end of the Robotic Parallelogram Arm provides independent translation and rotation of the precurved tube and optical laser fiber using two pneumatic stepper actuators (Fig.~\ref{fig:CTAUfig}). The precurved tube actuator independently translates and rotates the precurved tube, while the fiber actuator controls the optical fiber and is mounted on a carriage that translates with the precurved tube. Each actuator, 3D-printed in Nylon 12 via selective laser sintering, comprises a translational bellows, a rotational bellows, and two grippers (Fig.~\ref{fig:ActuatorsFig}A and ~\ref{fig:ActuatorsFig}B). The grippers function as mechanical clutches, selectively engaging and disengaging the tube from each bellows. When the fiber actuator's grippers are pressurized, the fiber passively rides along with the advancing precurved tube, maintaining a constant tip-to-tip relationship without coordinated multi-axis control. Releasing the grippers decouples the fiber for independent laser positioning.

These actuators retain the same stepwise operating principle as described in \cite{Peters2023}. Sequenced pressurization and exhaustion of the bellows and grippers produces bidirectional, inchworm-like motion. For example, pressurizing the linear bellows while its gripper is pressurized leads to translation in one direction. Alternatively, one can pressurize the linear bellows with its gripper exhausted. By subsequently pressurizing the gripper and exhausting the linear bellows, translation is achieved in the other direction. This same principle is applied to rotation. By following these sequences, one gripper remains pressurized and engaged at all times.

The primary design challenge of the CTAU was satisfying the severe space constraints imposed by the scanner bore and the anterior approach for MRI-guided, transforaminal access to the brain. Our original actuation unit \cite{Comber2016, comber2015dissertation} was too large to fit within the limited workspace. The actuators were therefore redesigned to reduce their overall length by nesting the grippers inside of the bellows, enabling them to occupy the same lengthwise space \cite{Peters2023}. We subsequently adopted a direct-drive architecture \cite{PetersHSMR2024} which further reduced the overall CTAU length and reduced the risk of buckling and torsional windup. This was achieved by eliminating intermediate tubes, known as transmission tubes, and grasping the precurved tube much closer to the tissue.

Eliminating transmission tubes also eliminated the mechanical advantage they provided. The actuators were therefore required to generate the forces and torques needed to overcome concentric tube friction and tissue interactions using small-diameter tubes that provide little mechanical advantage. The primary design advance presented here over \cite{PetersHSMR2024} is a new actuator with the force and torque capability to deploy the CTR under clinically relevant loading \cite{Peters2023}, while preserving the compact architecture that enables MRI-guided transforaminal access. 

To increase the force and torque capability of direct drive, we increased the frictional capacity of the tube--gripper interface. Since actuator size and operating pressure were constrained by the MRI workspace \cite{Grillo2023} and standard medical air ($345-379$ kPa) \cite{nfpa99_2024}, respectively, increasing effective bellows area and the coefficient of friction became the primary remaining design variables. Accordingly, the translational and rotational bellows were redesigned from circular \cite{PetersHSMR2024} to rectangular profiles (Fig. \ref{fig:ActuatorsFig}A and \ref{fig:ActuatorsFig}B), increasing the effective bellows area from 
$6 \text{ cm}^2$ \cite{PetersHSMR2024} to $40 \text{ cm}^2$ for the grippers, from $60 \text{ cm}^2$ \cite{PetersHSMR2024} to $240 \text{ cm}^2$ for the linear bellows, and from $26 \text{ cm}^2$ \cite{PetersHSMR2024} to $94 \text{ cm}^2$ for the rotary bellows. The gripper contact surfaces were also redesigned over those in \cite{PetersHSMR2024} to incorporate compliant TPU inserts to increase the coefficient of friction between the gripper and tube. Together, these design decisions prevented tube slip during CTR deployment in tissue-mimicking phantoms.

The final CTAU design demonstrates that compact pneumatic direct-drive architectures can overcome CTR intra-tube friction and tissue interaction forces while satisfying constraints on actuator length and operating pressure. With the precurved tube and laser fiber withdrawn inside the docking tube, the complete CTAU measures 150~mm $\times$ 110~mm $\times$ 270~mm (W $\times$ H $\times$ L) with a combined insertion stroke of 90 mm. This is a 73\% reduction in overall length compared to our original design \cite{Comber2016}, enabling it to fit and be manipulated within the scanner bore (Fig. \ref{fig:RoboticSystemOverviewCAD}). The actuators operate on a supply pressure of 345 kPa, consistent with hospital medical air \cite{nfpa99_2024}. Joint-level control is enabled by US Digital optical encoders for translation (EM2-0-2000-I, LIN-2000) and rotation (E16-4096). Collectively, these design choices produce a compact, MR-conditional actuation unit capable of deploying concentric tubes under clinically relevant loading while satisfying the constraints of MRI-guided transforaminal access.

\subsection{Robotic Parallelogram Arm (RPA)}

\begin{figure}[!t]
\centering
\includegraphics[width=\columnwidth]{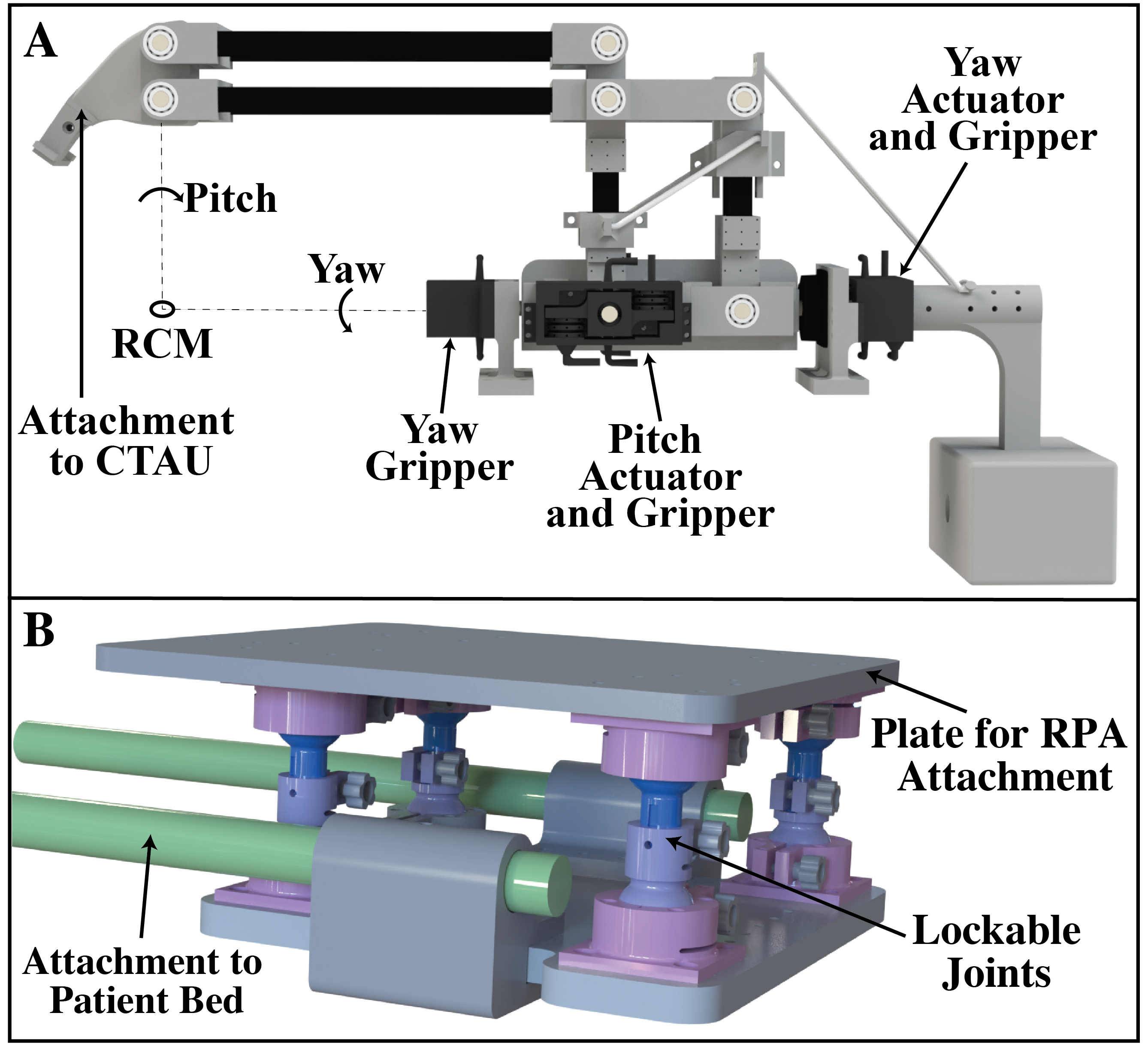}
\caption{(A) Robotic Parallelogram Arm (RPA) used to adjust the angle of the initial entry vector of the docking tube. The body of each actuator (shown in Fig. \ref{fig:ActuatorsFig}C) is affixed perpendicular to its axis of rotation in order to generate rotation in both the pitch and yaw directions. (B) A 6-DOF manual positioning platform aligns the RCM of the RPA to the foramen ovale. 
}
\label{fig:AimingArmFig}
\end{figure}

The Robotic Parallelogram Arm (RPA), shown in Fig. \ref{fig:AimingArmFig}A, orients and aims the CTAU, and consequently the docking tube, while maintaining a mechanical remote center-of-motion (RCM) at the docking tube tip \cite{grillo_6-dof_2025}. The RCM coincides with the foramen ovale, allowing the RPA to adjust the docking tube orientation through two rotational degrees of freedom (pitch and yaw) while maintaining a fixed tip position. The docking tube orientation established by the RPA defines the initial trajectory of the CTR. 

The RPA is implemented as a double-parallelogram mechanism, a well-established approach for achieving a mechanical RCM in surgical robots \cite{zhang_state_2024} (Fig.~\ref{fig:AimingArmFig}A). The RPA is constructed from square carbon-fiber links epoxied to 3D-printed interfaces to provide a lightweight, MR-conditional structure with high stiffness. Each rotational degree of freedom is actuated by a pneumatic stepper actuator consisting of a gripper mounted concentrically about the rotation axis and two radially arranged bellows (Fig.~\ref{fig:ActuatorsFig}C). This includes an additional gripper positioned on the opposing end of a common driveshaft. These rotary stepper actuators follow the same architecture as those used in the CTAU, but they incorporate larger grippers and bellows with greater effective area to generate the higher torques required to orient the CTAU and docking tube while resisting reaction forces generated during aiming. During operation, pressurized air first actuates the gripper to clamp the driveshaft and then inflates the rotational bellows to generate an incremental joint rotation. Repeating this sequence with alternating gripper engagement produces bidirectional stepwise motion analogous to that of the CTAU. One gripper remains engaged throughout the stepping sequence, preventing backdriving during gripper transitions. Closed-loop control of pitch and yaw is achieved using US Digital optical encoders (E3-10000-500-IE-H-M-3).

\subsection{Positioning Platform}

A passive, lockable 6-DOF positioning platform supports the RPA and CTAU while providing coarse alignment of the robot to the docking tube prior to robotic aiming. Since this alignment is performed only once during setup, manual adjustment provides a simpler, more compact, and stiffer alternative to a motorized positioning stage, thus reducing system complexity and preserving compatibility with the confined MRI environment. The platform mounts directly to the AtamA transfer board, allowing the robot to remain registered to the patient throughout transport between the fluoroscopy and MRI suites as described in Section~\ref{sec:workflow}, while integrating with the existing clinical LITT workflow.

To enable intuitive, simultaneous adjustment of all six degrees of freedom while maintaining high structural stiffness, the positioning platform employs a compact 4-SPS parallel architecture consisting of four manually adjustable 3-D printed legs connecting upper and lower acrylic plates (Fig.~\ref{fig:AimingArmFig}B). The surgeon can grasp the upper plate and reposition the entire robot in a single continuous motion, avoiding the sequential joint adjustments required by conventional serial positioning stages. The lower plate attaches to the AtamA board through fiberglass support rods behind the patient's head, and each leg contains a prismatic joint between two spherical joints. Once the RPA is aligned with the docking tube, integrated C-clamps and brass thumb screws lock all joints to form a rigid, scanner-compatible structure capable of supporting the cantilevered loads of the RPA and CTAU and resisting reaction forces generated during CTR aiming and deployment.

\subsection{CTR Fabrication}
\label{needledesign} 
For the experiments presented in this paper, we fabricated a patient-specific helical needle following the workflow described in Section~\ref{sec:workflow}. To evaluate the robot under the most demanding conditions, we selected a patient case requiring one of the highest-curvature trajectories from the image set in \cite{Granna2022}, since higher-curvature tubes generate larger reaction forces and require greater actuation force and torque. The chosen needle geometry had a target curvature of $\kappa=32.2~\mathrm{m}^{-1}$ and torsion of $\tau=81.1~\mathrm{m}^{-1}$.

A nitinol tube (90~mm long, 1.10~mm outer diameter, 0.93~mm inner diameter) was shape set in a custom-machined jig using a fluidized sand bath to match the desired helical geometry \cite{Granna2022}. Following fabrication, the tube was laser scanned using a Quantum X FaroArm\textsuperscript{\textregistered} laser scanner (FARO), and a helix was fit to the measured centerline according to \cite{taubin1991estimation}. The fabricated tube had a curvature of $\kappa=30.3~\mathrm{m}^{-1}$ and torsion of $\tau=81.0~\mathrm{m}^{-1}$, demonstrating close agreement with the intended design.

\begin{figure*}{}
\centering
\includegraphics[width=\columnwidth]{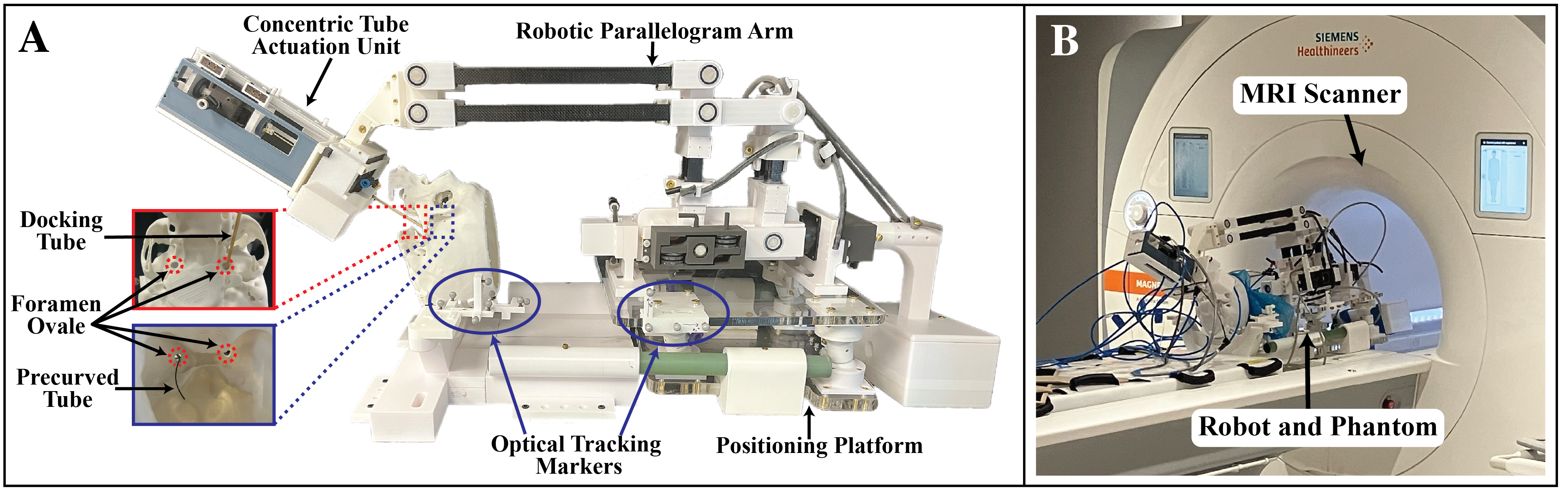}
\caption{Experimental setup with the assembled robotic system. (A) Robotic hardware for optical tracker experiments. (B) Robotic system inside the MRI scanner. }
\label{fig:ExpSetup}
\end{figure*}

\section{Results}\label{sec3}

\subsection{Free-Space Targeting Experimental Procedure and Results}

\begin{figure}[h]
\centering
\includegraphics[width=.75\columnwidth]{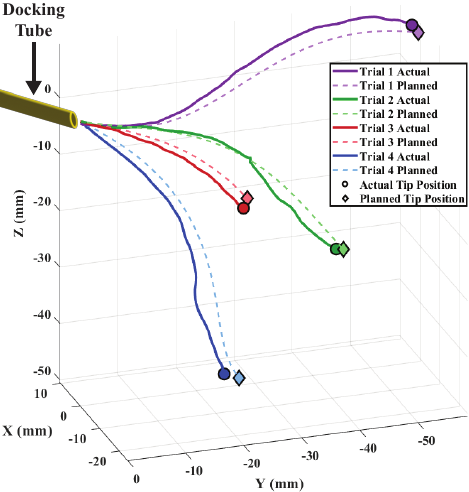}
\caption{Free-space targeting results for four desired trajectories. The final tip errors were 1.70 mm, 1.75 mm, 1.18 mm, and 2.47 mm for Trials 1-4, respectively.} 
\label{fig:FreeSpace_SysTargeting}
\end{figure}

We conducted initial experiments in free space using an OptiTrack Prime 13 optical tracking system (NaturalPoint, Inc) to measure CTR deployment (Fig. \ref{fig:ExpSetup}A). The robot was commanded to deploy the precurved tube to four separate targets using follow-the-leader deployment \cite{gilbert2015concentric}, with an optically tracked spherical marker affixed to the tube tip. The planned and actual paths of the trials are shown in Fig.\ \ref{fig:FreeSpace_SysTargeting} with tip placement errors of 1.70, 1.75, 1.18 and 2.47 mm. Note that the range of the placement errors is comparable to the free-space performance reported for prior MRI-guided neurosurgical robots (see e.g.\ \cite{li2020fully, li2014robotic,chen2019mr, gunderman2023non}).

\subsection{MRI-Guided Targeting Experimental Procedure and Results}

To validate the operation of the entire system and measure the ultimate end-goal of hippocampal cannulation, we performed transforaminal deployment experiments in the MRI in tissue-mimicking brain phantoms. The experiments were performed using a 3T Siemens MAGNETOM Vida scanner (Siemens Healthineers), which features a 70 cm bore, at University Hospital Cleveland Medical Center (Cleveland, OH) (Fig. \ref{fig:ExpSetup}B). 

Our phantom consisted of a skull model that was 3D printed from polylactic acid (PLA) using a Bambu X1 Carbon 3D printer (Bambu Lab). The skull was generated from a patient case reported in \cite{Granna2022}, using the same patient-specific parameters used to fabricate the needle described in Section \ref{needledesign}. The skull was filled with brain-tissue-mimicking agarose gel doped with copper sulfate (1\% weight/volume agar, 0.1g CuSO$_4$/100 mL water) for MRI contrast \cite{mitchell1986agarose}. MRI-visible markers (Beekley MR-SPOT\textsuperscript{\textregistered}\,122) were affixed at known locations on both the robot and phantom, enabling robot-image registration for MRI-guided trajectory following. The phantom and robot were mounted to the AtamA board before the positioning stage was adjusted to seat the docking tube tip in the left foramen ovale. The positioning stage was subsequently locked, and MRI transceiver coils (Siemens Medical Solutions USA, Inc. UltraFlex Large 18) were secured around the robot and phantom. A preoperative T1-weighted MPRAGE (TE: 4 ms, TR: 8.6 ms, Flip Angle: 20$\degree$, Voxels: 1~mm $\times$ 1~mm $\times$ 1~mm, FOV: 250~mm $\times$ 250~mm $\times$ 208~mm) volumetric scan was acquired, and the MRI-visible markers were localized in 3D Slicer (version 5.8.1) to register image and robot frames via a least-squares optimization \cite{3dslicerwebsite, fedorov_3d_2012}.

\begin{figure}[H]
\centering
\includegraphics[width=0.75\columnwidth]{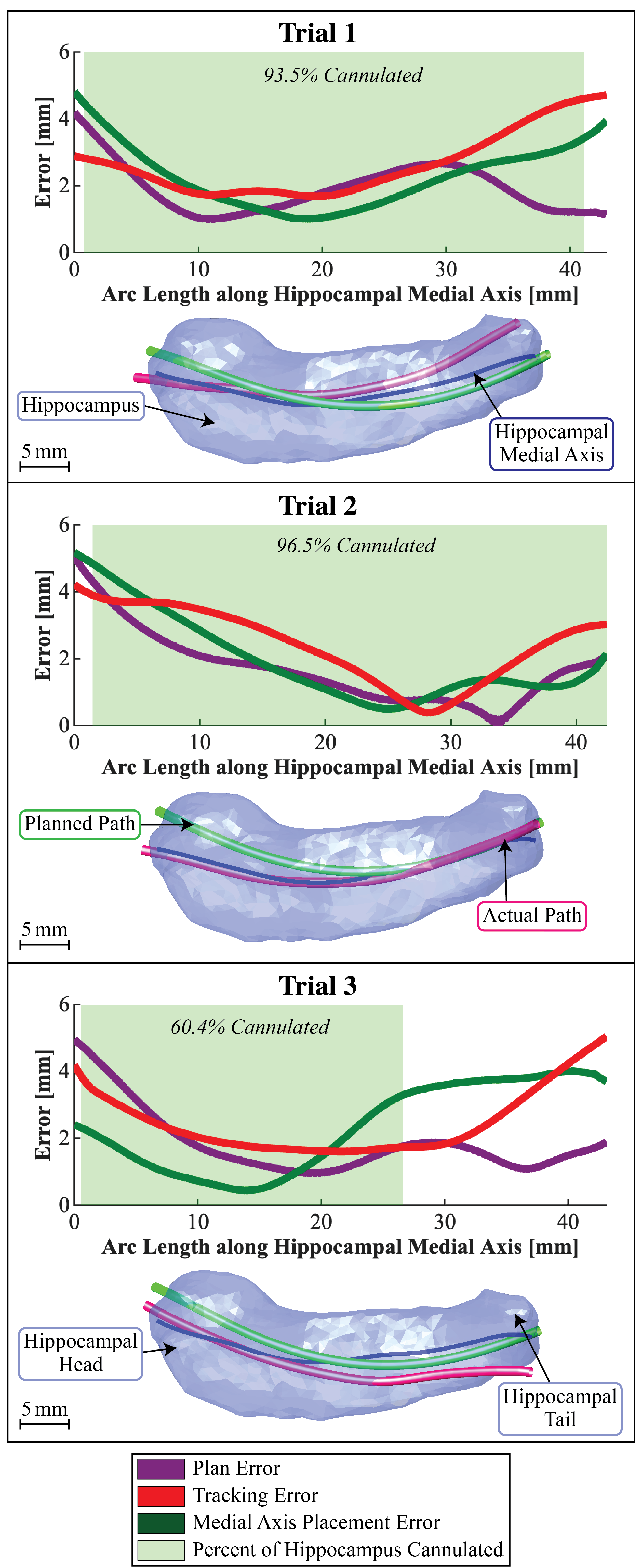}
\caption{MRI‑guided phantom evaluation of the curved, transforaminal approach (n=3). For each trial, four spatial performance metrics are plotted versus arc length along the hippocampal medial axis (s = 0 at the head, increasing toward the tail; total length $\approx$ 40 mm). Corresponding 3D renderings depict the segmented hippocampus (semi‑transparent blue), its medial axis (dark blue), the planned trajectory (green), and the actual trajectory (magenta). Pale green shading marks the percent of the hippocampus cannulated by the actual path.
} 

\label{fig:ErrorPlot}
\end{figure}

The orientation of the docking tube was adjusted by the RPA according to the plan from \cite{Granna2022} for this patient case, and the CTR was deployed transforaminally into the brain phantom with the goal of following the medial axis of the hippocampus. A final volumetric, T1-weighted MPRAGE scan was acquired, and the signal void around the CTR was segmented in 3D Slicer. The CTR trajectory was approximated using the centerline of this segmentation, which was used to calculate the performance metrics. This protocol was repeated for three trials. While the same hippocampus is targeted in all cases, the optimal plan varied across trials because the docking tube is placed into the foramen ovale of a fresh phantom for each trial, which introduces small tube tip positioning variations relative to target anatomy. The results of these experiments are shown in  Fig.~\ref{fig:ErrorPlot} and summarized in Table~\ref{tab:CaseResultsTable}.

The Percent of Hippocampus Cannulated quantifies how much of the hippocampus contains the CTR. Defined in the clinical literature for straight insertions, this metric is hypothesized to enable more complete treatment and hence better outcomes, thus surgeons seek to maximize it \cite{wu2015effects}. We achieved cannulation percentages of 93.5\%, 96.5\%, and 60.4\% in Trials 1–3 (Table \ref{tab:CaseResultsTable}), which exceed the 50-60\% reported for traditional LITT \cite{wu2015effects,vakharia_automated_2018,wu2014extraventricular,kang2016laser}. Cannulated portions of the insertion are also illustrated with green shading in Fig.\ \ref{fig:ErrorPlot}.

To quantify the accuracy of the CTR placement, we evaluated the agreement between the planned trajectory, the actual trajectory, and the hippocampal medial axis. The Plan Error is the distance between the hippocampal medial axis and the planned CTR trajectory in 3D Slicer, measuring how well the CTR shape and pose in preoperative planning aligned with the target anatomy. Since the medial axis is approximated as a helix and the CTR has manufacturing tolerances, this error is generally nonzero. The Tracking Error is the distance between the planned and actual trajectories, where the actual trajectory is derived from the signal void around the CTR as mentioned previously. The Medial Axis Placement Error is the distance between the hippocampal medial axis and the actual CTR trajectory. Across all three trials, the mean Plan Error was 1.80 mm, the mean Tracking Error was 2.78 mm, and the mean Medial Axis Placement Error was 2.18 mm (Table \ref{tab:CaseResultsTable}) (Fig. \ref{fig:ErrorPlot}).

\begin{table*}[t]
\caption{Performance Metrics along the Hippocampal Arc Length for MRI-Guided Targeting Experiments}
\label{tab:CaseResultsTable}
\centering

\begin{tabular}{c|c|cc|cc|cc}
\hline
& \multicolumn{7}{c}{\textbf{Performance Metric}} \\
\hline

\multirow{2}{*}{\textbf{Trial}}
&
\multirow{2}{*}[1ex]{
  \begin{tabular}[c]{@{}c@{}}
  \textbf{Percent of}\\
  \textbf{Hippocampus}\\
  \textbf{Cannulated (\%)}
  \end{tabular}
}
&
\multicolumn{2}{c|}{
  \begin{tabular}[c]{@{}c@{}}
  \textbf{Plan}\\
  \textbf{Error (mm)}
  \end{tabular}
}
&
\multicolumn{2}{c|}{
  \begin{tabular}[c]{@{}c@{}}
  \textbf{Tracking}\\
  \textbf{Error (mm)}
  \end{tabular}
}
&
\multicolumn{2}{c}{
  \begin{tabular}[c]{@{}c@{}}
  \textbf{Medial Axis}\\
  \textbf{Placement}\\
  \textbf{Error (mm)}
  \end{tabular}
}
\\

&
&
\textbf{Mean} & \textbf{Max}
& \textbf{Mean} & \textbf{Max}
& \textbf{Mean} & \textbf{Max}
\\
\hline

\textbf{1}
& 93.5
& 1.91 & 4.19
& 2.90 & 4.85
& 2.28 & 4.81 \\

\textbf{2}
& 96.5
& 1.67 & 5.04
& 2.64 & 5.43
& 1.95 & 5.16 \\

\textbf{3}
& 60.4
& 1.83 & 4.95
& 2.79 & 5.73
& 2.31 & 4.01 \\

\hline

\textbf{Mean}
& \textbf{83.5}
& \textbf{1.80} & 4.73
& \textbf{2.78} & 5.34
& \textbf{2.18} & 4.66 \\

\hline
\end{tabular}
\end{table*}

\section{Discussion}\label{sec4}

In this paper, we described a new robotic system for curvilinear transforaminal LITT. The transforaminal approach may reduce invasiveness by eliminating burr holes, and the curvilinear trajectory could provide better conformity to the curved shape of the hippocampus. Overall, our experiments demonstrated improved cannulation percentages of 93.5\%, 96.5\%, and 60.4\% in comparison to 50-60\% cannulation percentages in the clinical literature. Additionally, the mean Tracking Error of 2.78 mm across all three trials is comparable to the accuracy reported for previous surgical robots \cite{li2014robotic, li2020fully, monfaredi2024automatic} and is consistent with prior results for MRI-guided CTRs \cite{su2016concentric}, which reported tip errors of 2.18 mm in gelatin and 4.64 mm in ex vivo liver tissue.

While the clinically required accuracy of needle placement for LITT is not yet well established, surgeons use cannulation as an indirect measure of the ability to ablate epileptogenic tissue, ideally with the thermal lesion encompassing the entire epileptogenic region \cite{youngerman2020magnetic}. In current LITT, infrared light is absorbed, generating heat that spreads to surrounding tissue through thermal conduction \cite{kaub2023comparison,majdabadi2014analysis}. This produces a lesion that extends beyond the physical fiber trajectory with reported lesion dimensions on the order of 10-20 mm, depending on the power applied, treatment duration, fiber design, and tissue conditions \cite{ahrar2010preclinical}. The resulting geometry is further shaped by heat-sink effects from surrounding brain tissue, vasculature, and cerebrospinal fluid. In this context, successive LITT lesions along the actual CTR trajectories executed in this work could plausibly compensate for the millimeter-scale Medial Axis Placement Error reported. 

Furthermore, we believe a highly promising direction for future research is directional ablation. Commercial directional ablation systems have already reached clinical practice, including the Monteris Medical NeuroBlate system \cite{Sloan2013} and the Profound Medical TULSA-PRO system \cite{fung2024mr}. Research efforts have also investigated directional ablation using microwave \cite{mcwilliams2015directional}, focused ultrasound \cite{chopra2005method,tang2007conformal,ghoshal2013ex,scott2013interstitial,boctor2010precisely,burdette2010acusitt,carias2014evaluation}, and laser systems \cite{ngo2006side,nguyen2017fabrication,giglio2022reciprocating,george2011performance}. By preferentially directing thermal energy, these systems can shape the ablation zone to better conform to the target anatomy, while potentially compensating for trajectory placement errors. Directional ablation could substantially relax the CTR placement accuracy requirements and may enable complete conformal ablation throughout the cannulated regions of the hippocampus. 

In summary, we have presented an MRI-guided robotic system that increases hippocampal cannulation beyond what is achieved clinically. The system integrates a Concentric Tube Actuation Unit, a Robotic Parallelogram Arm, and a manual positioning platform to accomplish curved trajectories under MRI guidance. In addition to laser integration, future work will also include cadaveric studies to assess accuracy and the MRI-guided clinical workflow. In these studies, MR thermometry will be used to monitor treatment and confirm that the intended thermal dose is delivered to the target brain region. Ablation studies will then be performed to determine how increased cannulation and curved trajectories translate into improved volumetric ablation coverage. Collectively, our system and the above future research areas advance efforts to improve seizure control outcomes while reducing invasiveness through more complete, single-trajectory ablation coverage of the hippocampus.

\backmatter

\bmhead{Acknowledgements}

Research reported in this publication was supported by the National Institute of Biomedical Imaging and Bioengineering (NIBIB) of the National Institutes of Health under Award No. R01NS120518. A total of \$2,013,213, or 100\%, of the research project was financed with federal funds. The work reported in this publication represents a subset of the research conducted under this project. Daniel Esser wishes to acknowledge separate support from a fellowship from the Natural Sciences and Engineering Research Council (NSERC), reference number 521537544, that partially supported him during his graduate studies. The authors would also like to thank the Digital Fabrication Lab at Vanderbilt University for their manufacturing support. The content is solely the responsibility of the authors and does not necessarily represent the official views of the National Institutes of Health.

\bmhead{Conflict of Interest}
The authors have no conflicts of interest to report.

\bibliography{systempaper}

\end{document}